\documentclass{article}
\usepackage[utf8]{inputenc}
\usepackage{authblk}
\usepackage{setspace}
\usepackage[margin=1.25in]{geometry}
\usepackage{graphicx}
\graphicspath{ {./figures/} }
\usepackage{subcaption}
\usepackage{amsmath}

\usepackage[font=small,labelsep=period]{caption} 

\usepackage{multirow}
\usepackage{amssymb}% http://ctan.org/pkg/amssymb
\usepackage{pifont}% http://ctan.org/pkg/pifont
\newcommand{\cmark}{\ding{51}}%
\newcommand{\etal}{\textit{et al.}}
\usepackage{graphicx}
\usepackage{tikz}
\usetikzlibrary{arrows.meta}
\usepackage{booktabs}
\usepackage{float}
\usepackage[section]{placeins}
\usepackage{enumitem}
\makeatletter
\newcommand\figcaption{\def\@captype{figure}\caption}
\newcommand\tabcaption{\def\@captype{table}\caption}
\makeatother

\usepackage[style=nejm, 
citestyle=numeric-comp,
sorting=none]{biblatex}
\title{Multi-Scale Cross Contrastive Learning for Semi-Supervised Medical Image Segmentation}

\author[1]{Qianying Liu}
\author[2]{Xiao Gu}
\author[1]{Paul Henderson}
\author[1]{Fani Deligianni}

\affil[1]{School of Computing Science, University of Glasgow, Glasgow, UK.}
\affil[2]{School of Computing Science, Imperial College London, London, UK.}

\date{}

\begin{document}

\maketitle

%%%%%% Abstract %%%%%%
\begin{abstract}
Semi-supervised learning has demonstrated great potential in medical image segmentation by utilizing knowledge from unlabeled data.
However, most existing approaches do not explicitly capture high-level semantic relations between distant regions, which limits their performance.
In this paper, we focus on representation learning for semi-supervised learning, by developing a novel Multi-Scale Cross Supervised Contrastive Learning (MCSC) framework, to segment structures in medical images. We jointly train CNN and Transformer models, regularising their features to be semantically consistent across different scales. Our approach contrasts multi-scale features based on ground-truth and cross-predicted labels, in order to extract robust feature representations that reflect intra- and inter-slice relationships across the whole dataset. To tackle class imbalance, we take into account the prevalence of each class to guide contrastive learning and ensure that features adequately capture infrequent classes.
Extensive experiments on two multi-structure medical segmentation datasets demonstrate the effectiveness of MCSC. It not only outperforms state-of-the-art semi-supervised methods by more than 3.0$\%$ in Dice, but also greatly reduces the performance gap with fully supervised methods.
\end{abstract}

%%%%%% Main Text %%%%%%

\section{Introduction}  

% Draft of Introduction by XG
%\input{introduction}

Image segmentation serves as a fundamental process in medical image analysis by delineating organ structures and allowing the quantification of their shape and size, thus providing essential information for clinical diagnostics, treatment planning, and patient monitoring \cite{chen2020deep,shen2017deep}. Deep learning approaches have achieved great successes in medical image segmentation in recent years; however, such techniques hinge upon the availability of large-scale and accurately annotated datasets \cite{tajbakhsh2020embracing}.
In the medical domain, such datasets require prohibitive time, cost, and expertise to obtain. To mitigate this issue, \textit{semi-supervised} learning (SSL) aims to minimize the annotation efforts by training with both labelled and unlabelled data \cite{Luo2022,Zhong2021,Ouali2020}. 

Several strategies have been proposed for SSL in medical image segmentation. These include iterative pseudo-labeling \cite{Seibold2022}, regularization strategies \cite{Luo2022,Chen2021a,Ouali2020,Xia2020,Wang2022}, as well as leveraging domain-specific prior knowledge such as anatomical information \cite{Yang2022Self}. Typically pseudo-labeling iteratively generates approximate segmentation masks for unlabeled data. 
Integrating these pseudo annotations with ground truth labels for model updates necessitates a meticulously designed approach, which remains an open problem. Differently, several regularization approaches forgo this process, by enforcing prediction consistency over different data transformations \cite{Chen2021a,Xia2020}, different model architectures \cite{Luo2022,Ouali2020}, or different tasks \cite{Wang2022}. In particular, recent works \cite{Luo2022} have investigated the possibility of making use of two advanced segmentation backbones, e.g., CNN and Transformer, for cross-teaching SSL. 

Although these methods are promising, their performance is significantly weaker than fully supervised approaches and thus their practical application in medical image segmentation is limited \cite{tarvainen2017mean,qiao2018deep,yu2019uncertainty}. To alleviate this issue, \textit{contrastive learning} has been extensively utilized to facilitate robust feature learning. It functions by encouraging feature similarity of positive pairs, as well as dissimilarity of negative pairs. Positive pairs may be defined in a self-supervised manner as different augmentations of the same instance \cite{Chen2021} or in a supervised manner based on the actual label \cite{Khosla2020}. 
%\textcolor{red}{(self-supervised contrastive on global and local level)}
In SSL, pioneering works \cite{Wu2022Cross,chaitanya2020contrastive} have made efforts towards directly applying contrastive learning on unlabelled data, by performing global-level image contrast for training. However, this strategy is mostly suited for classification tasks, since it extracts global representations that ignore detailed pixel-level information. To accurately delineate organ boundaries, a local contrastive strategy is required to enable predictions at a pixel level \cite{Zhong2021,chaitanya2020contrastive,Wang2021a}. In particular, for image segmentation that inherently relies on dense-wise prediction, Chaitanya \etal{} highlighted the importance of complementing the global image-level contrast with local pixel-level contrast \cite{chaitanya2020contrastive}. 

%\textcolor{red}{(supervised contrastive on local level)}
Since self-supervised contrastive learning normally select augmented views of the same sample data point as positive pairs \cite{Chen2021}, without prior knowledge of the actual class label and its prevalence, it is prone to a substantial number of false negative pairs, particularly when dealing with class-imbalanced medical imaging segmentation datasets \cite{li2020analyzing}.
To mitigate the false negative predictions resulting from self-supervised local contrastive learning, existing works have investigated supervised local contrastive learning \cite{Hu2021, Chaitanya2023}. 
Pioneering works \cite{zhao2023rcps, Wu2022Cross} applied supervised contrastive learning only on unlabelled data based on conventional iterative pseudo annotation.  
Some studies \cite{Hu2021} also attempted to apply supervised local contrastive loss on labelled data exclusively, whilst performing self-supervised training for unlabelled data. However, the discrepancy in positive/negative definitions leads to divergent optimization objectives, which may yield suboptimal performance. 

% [insert work  \cite{zhao2023rcps, Wu2022Cross) } do local supervised on unlabeld data guided by psudo label first (most of work), then followed by this ref 13 on labeled (only one )]Some studies applied supervised local contrastive loss on labelled data exclusively, whilst performing self-supervised training for unlabelled data. The discrepancy in positive/negative definitions leads to divergent optimization objectives, which may yield suboptimal performance \cite{Hu2021}. 

%In this sense \pmh{??}, the pseudo label introduced in cross-teaching semi-supervised learning framework, brings about the possibility to explore a unified supervised contrastive learning for both labelled and unlabelled data. Nevertheless, the challenges posed by the inherent class imbalance, as well as the limited sample size remains unresolved, constraining the network from generating unbiased and generalizable representations.  
We propose a novel multi-scale cross contrastive learning framework for semi-supervised medical image segmentation. Both labelled and unlabelled data are integrated seamlessly via cross pseudo supervision and balanced, local contrastive learning across features maps that span multiple spatial scales. Our main contributions are three-fold: 
\begin{itemize}[leftmargin=10pt,itemsep=0.5pt,topsep=2pt]
  \item We introduce \textbf{a novel SSL framework} that combines the benefits of cross-teaching with a proposed local contrastive learning. This enhances training stability, and beyond this, ensures semantic consistency in both the output prediction and the feature level.
  \item We develop the first \textbf{local contrastive framework} defined over \textbf{multi-scale feature maps}, which accounts for over-locality and over-fitting typical of pixel-level contrast. This benefits from seamlessly unifying pseudo-labels and ground truth via cross-teaching.
  \item We incorporate a \textbf{balanced contrastive} loss to enforce \textbf{unbiased representation learning} in semi-supervised medical image segmentation. 
  This tackles the significant imbalance issue for both pseudo label prediction, and  
  the concurrent supervised training based on imbalanced (pseudo) labels. 
 % by averaging the gradient contributions of all negatives of each class for every mini-batch. This ensures that contrastive learning extracts unbiased feature representations.
\end{itemize}

We evaluate our proposed methodology on two challenging benchmarks of radiological scans: multi-structure MRI segmentation on ACDC \cite{bernard2018deep}, and multi-organ CT segmentation on Synapse \cite{landman2015miccai}.
Our approach not only significantly outperforms state-of-the-art SSL methods, but also closes the gap between fully supervised approaches with just a small fraction of labelled data. With just $10\%$ labelled data, it achieves remarkable improvement in Hausdorff Distance (HD) from 8.0 to 2.3mm.
Our method is also more resilient to the reduction of labelled cases, achieving around $10\%$ improvement in Dice Coefficient (DSC) when labelled data are reduced from $10\%$ to $5\%$ in ACDC and from $20\%$ to $10\%$ in Synapse.

\vspace{-5pt}
\section{Related Work} 
\vspace{-5pt}

\paragraph{Consistency Regularization in Semi-Supervised Medical Image Segmentation.}
Semi-supervised learning has gained popularity in medical image segmentation due to its effectiveness in handling scenarios with limited annotations \cite{peng2020deep,bortsova2019semi,Luo2022,chen2020deep}. Among various approaches, enforcing prediction consistency has emerged as a crucial regularization strategy for extracting and leveraging knowledge from unlabelled data. Such regularization can be based on predictions from different augmentations \cite{bortsova2019semi,peng2020deep}, different architectures \cite{Luo2022}, or tasks \cite{Wang2022}. For instance, inspired by the fact that the predicted mask should undergo the same spatial transformations as the input images,  
Bortsova \etal{} \cite{bortsova2019semi} developed a transformation consistency based semi-supervised framework. Peng \etal{} \cite{peng2020deep} sought to attain prediction similarity from a batch of co-trained models with identical architectures, while adversarially preserving each model's diversity. 
Recent works \cite{Luo2022} has taken advantage of the advanced U-Net and Transformer, and aimed to achieve the prediction consistency from networks. However, the medical image datasets are typically imbalanced, which poses great challenges in learning unbiased predictions with limited annotations \cite{li2020analyzing}. Tackling such issue in consistency settings for unlabelled data remains an open problem. Furthermore, existing works primarily focus on prediction consistency at the output level \cite{Luo2022}, neglecting the pursuit of discriminative feature representations for both labelled and unlabelled data.

\paragraph{Contrastive Learning in Medical Image Segmentation.}
Contrastive learning has contributed to most successful self-supervised visual representation methods \cite{Chen2020,Grill2020,Tian2021,He2020}. The core idea is to promote the similarity of positive image pairs, whilst distinguishing negative pairs. To tailor for the needs of dense-wise downstream segmentation task, pixel-wise self-supervised contrastive learning has been introduced recently \cite{Xie2021, Wang2021}. Recent research has also found that integrating the contrastive loss in both global and local levels, can enhance performance \cite{chaitanya2020contrastive}.  
In the realm of natural images, there has been a growing interest in merging semi-supervised learning with contrastive learning, resulting in a one-stage, end-to-end model that forgoes unsupervised pretraining \cite{Zhong2021,Yang2022}. This approach has recently been adopted in the medical domain for segmentation tasks \cite{Chaitanya2023,Hu2021,zhao2023rcps,Wu2022Cross}. However, as discussed in the Introduction section, existing combinations of contrastive learning and semi-supervised learning do not fully address the inherent challenges posed by size-limited and data-imbalanced medical datasets, thus lacking generality. The question of how to effectively integrate contrastive learning for medical image segmentation remains open.

\vspace{-5pt}
\section{Method}
\vspace{-5pt}

\begin{figure}[th]
\begin{center}
\includegraphics[width=0.9\linewidth]{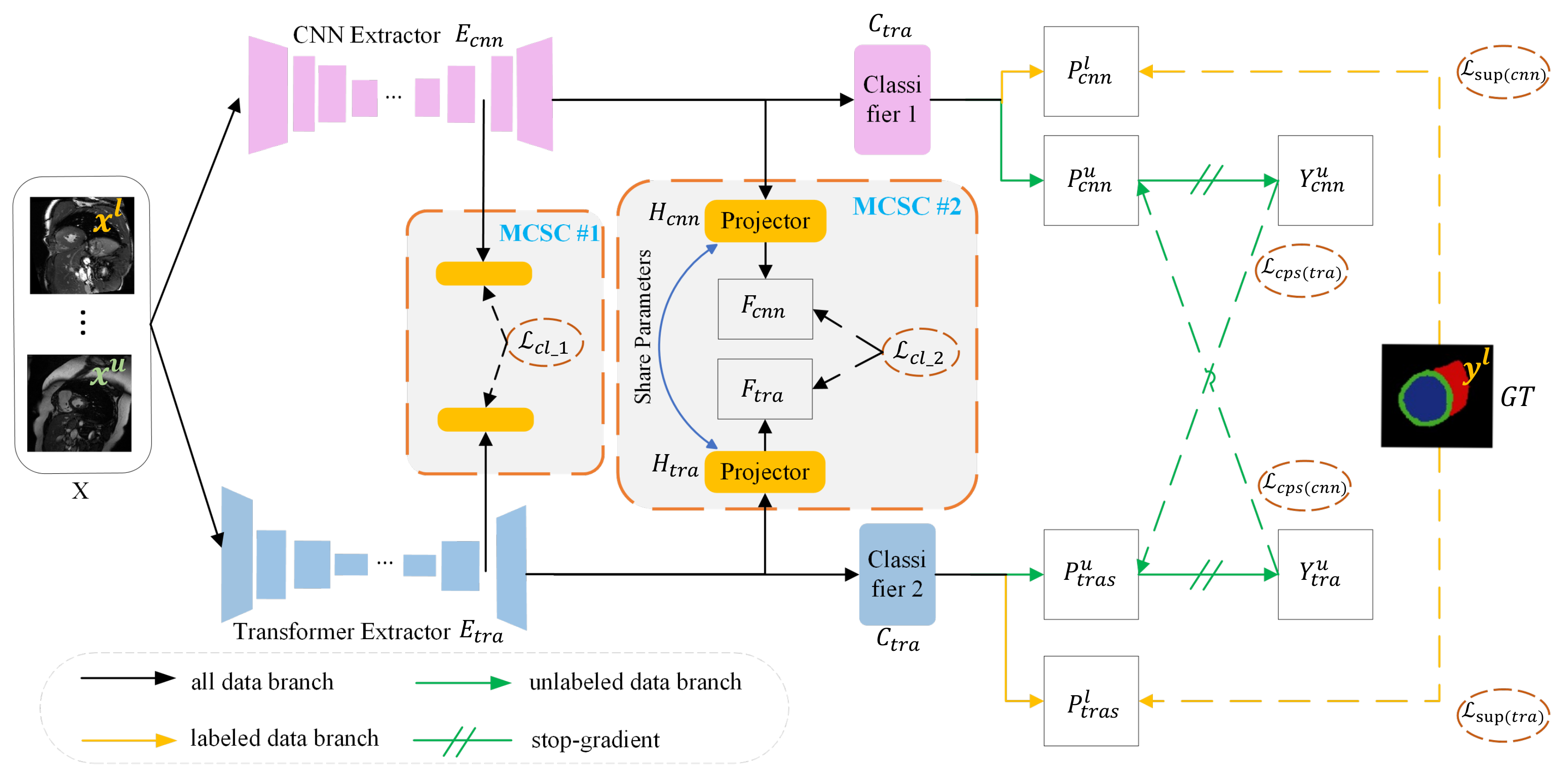}
    \end{center}
\vspace{-20pt}
\caption{The overall architecture of our MCSC framework for semi-supervised segmentation.
Two networks, a CNN (\textcolor[RGB]{240,185,238}{pink}) and Transformer (\textcolor[RGB]{159,194,224}{blue}), with complementary inductive biases, learn together.
When training on unlabelled data, each network generates pseudo labels for the other.
These labels are used to define a pseudo supervision loss and a novel local contrastive loss that improves the quality of representations learnt by the models.
}
\vspace{-8pt}
\label{fig:framework}
\end{figure}

%We consider multi-class segmentation in the semi-supervised setting.

We adopt a student--student framework based on \cite{Luo2022}, with cross-teaching between a CNN-based U-Net and a Transformer-based U-Net.
This leverages the advantages of convolution-based and Transformer-based segmentation networks for learning local semantic information and long-range dependencies, and enables the two models to achieve consistency on segmentation prediction. However, this framework has some limitations: (i) it only focuses on the prediction consistency on each image slice at output level; (ii) it ignores the dissimilarity and similarity among different and same segmentation categories across the whole dataset. To overcome this, we propose a Multi-Scale Cross Supervised Contrastive Learning (MCSC) framework to pull closer the features of the same category and push away the features of different categories from both networks. It not only ensures the consistency of two models on the feature and output level, but also enhances the distinguishability of features in different categories, thereby improving the segmentation performance. 
We illustrate the overall architecture of our framework in Figure \ref{fig:framework}.
The branch of CNN or Transformer includes a feature extractor $E_{*}(\cdot)$, a segmentation head $C_{*}(\cdot)$, and two feature space projectors $H_{*}(\cdot)$. Both branches only share the parameters of the last layer in the feature space projectors.

%In this way, the models are more likely to good at understanding of local semantic features to capture more subtle organ boundaries (merits of convolution) and also of global feature dependencies like relationships among multiple organs (merits of Transformer).
% as it achieves good representation learning on unlabeled data by takes advantage of locality of convolutions and long-range modeling ability of Transformer. 
% Each network provides pseudo-labels to the other, enabling cross-pseudo-label supervision and cross-contrastive learning simultaneously.
Given a training dataset consisting of a small labelled subset $D_{l} = \{x_{i}^l,y_{i}^l\}_{i=1}^K$ and a large unlabelled set $D_{u} = \{x_{j}^u\}_{j=1}^M$, where $M \gg K$, the input to our model is a minibatch $X = X^l \cup X^u$ including labelled images and unlabelled images. The minibatch $X$ is first fed into the CNN-based and Transformer-based networks to obtain their feature representations and segmentation logits. In the semi-supervised setting, we employ the following supervision losses for training:
(i) on the output level, we calculate the \textit{supervision loss} $\mathcal{L}_{sup}$ (\textcolor[RGB]{255, 207, 0}{yellow dashed lines} in Figure \ref{fig:framework}) between the segmentation predictions and the limited labelled data, as well as the \textit{cross pseudo supervision loss} $\mathcal{L}_{cps}$ (\textcolor[RGB]{15, 178, 89}{green dashed lines} in Figure \ref{fig:framework}) between the segmentation predictions and the pseudo labels from the CNN-based U-Net or the Transformer-based U-Net in a cross teaching manner on the output level (Section~\ref{subsec:cross-pseudo-sup}); 
(ii) on the feature level, we employ the proposed multi-scale cross contrastive loss $\mathcal{L}_{cl}$ (black dashed lines in Figure \ref{fig:framework})
to enhance feature consistency of the same segmentation category and feature distinguishability of the different segmentation categories across the whole dataset (labelled and unlabelled) guided by cross pseudo-supervision (Section~\ref{subsec:cscl}).
 %The \textit{supervised learning branch} (\textcolor[RGB]{255, 207, 0}{
%\begin{tikzpicture}[baseline=-\the\dimexpr\fontdimen22\textfont2\relax]
%\draw[-latex, line width=2pt] (0,0) -- (1.,0);
%\end{tikzpicture}}
%in \figureautorefname{}~\ref{fig:framework}) calculates $\mathcal{L}_{sup}$, which includes the standard cross-entropy and dice losses, between classifier output predictions and ground truth on labeled data.

%for unsupervised data (\textcolor[RGB]{15, 178, 89} {\begin{tikzpicture}[baseline=-\the\dimexpr\fontdimen22\textfont2\relax]\draw[-latex, line width=2pt] (0,0) -- (1.,0);
%\end{tikzpicture}} in \figureautorefname{}~\ref{fig:framework}), we use \textit{cross pseudo supervision} $\mathcal{L}_{cps}$ to encourage predictions from the CNN and Transformer to be consistent (Section~\ref{subsec:cross-pseudo-sup}).

%\textit{Cross supervised contrastive learning in multi-scale} (\textcolor{black}{
%\begin{tikzpicture}[baseline=-\the\dimexpr\fontdimen22\textfont2\relax]
%\draw[-latex, line width=2pt] (0,0.) -- (1.,0.) ; 
%\end{tikzpicture}} operates on features of all data to exploit feature structures across the whole dataset guided by mutual pseudo-supervision (Section~\ref{subsec:cscl}). 

\begin{figure}[tb!]
\begin{center}
\includegraphics[width=0.9\linewidth]{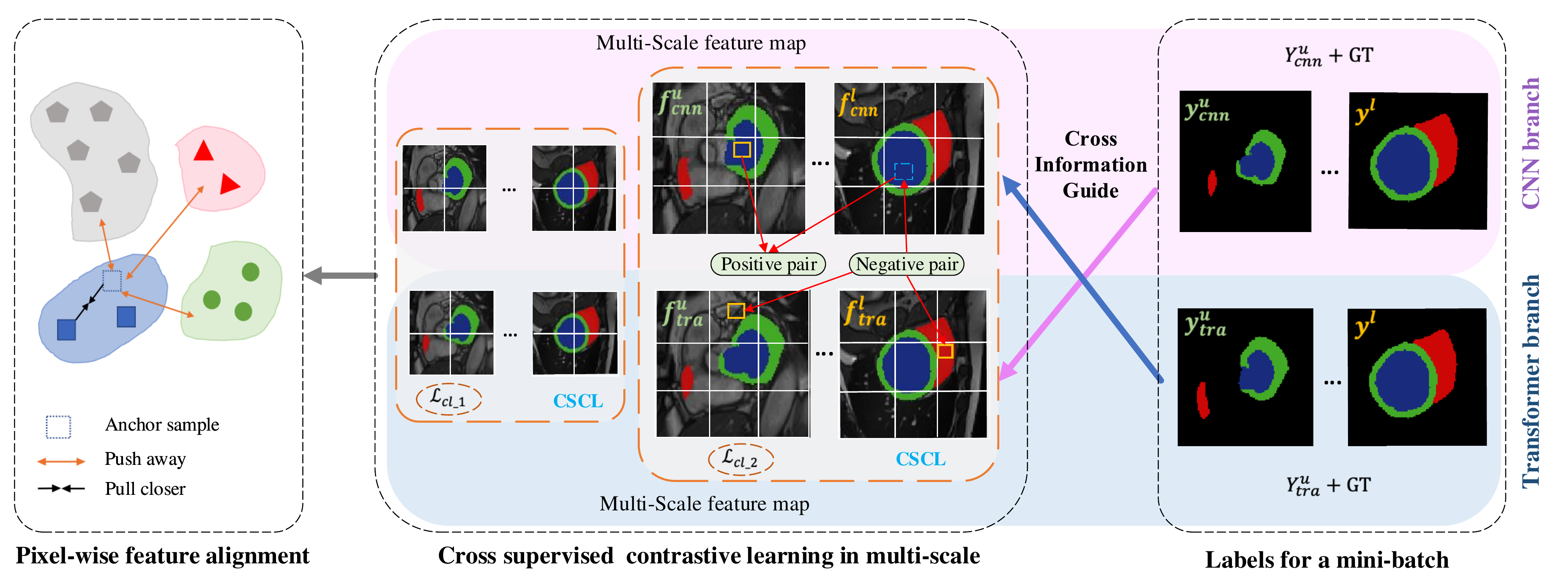}
    \end{center}
\vspace{-15pt}
\caption{Multi-scale cross supervised contrastive learning. Pseudo labels from cross-teaching (\textbf{right}) are combined with ground-truth labels where available, and used to define a local contrastive loss over features of different scales (\textbf{middle}, \textcolor[RGB]{238,139,75}{orange dashed boxes}). This contrastive pairs of pixels drawn from either the same or different slices; for efficiency it is defined over patches. Features of pixels of the same (pseudo-) class are pulled together (\textbf{left}), while those of different classes are pushed apart.}
\label{fig:contrastive}
\vspace{-10pt}
\end{figure}

\subsection{Cross Pseudo Supervision}
\label{subsec:cross-pseudo-sup}

The CNN and Transformer networks teach each other using the unlabelled data, through a cross pseudo supervision loss $\mathcal{L}_{cps}$ \cite{Luo2022, Chen2021a}. This regularises their respective predictions to be consistent with each other. Specifically, the predictions made by the CNN become pseudo labels that supervise the Transformer, and vice-versa.
The unlabelled images $X^{u}$ are fed into the feature extractors $E_{*}(\cdot)$ and classifier heads $C_{*}(\cdot)$ of the two models respectively, to get class probability maps $P^u_{*} = \mathrm{softmax} \{C_{*}(E_{*}(X^u)) \}$, and pseudo one-hot label map $Y^u_{*} = \mathrm{argmax}\,( P^u_{*})$, where $*$ denotes the CNN or Transformer branch.
We then define two consistency loss terms: $\mathcal{L}_{cps (cnn)} $ uses the Transformer's pseudo-labels to supervise the CNN, and $\mathcal{L}_{cps (tra)} $ the reverse; these are given by:
%
% \begin{align}
% \label{CPS}
%  \mathcal{L}_{cps (cnn)} &= \mathcal{L}_{dice}( P^u_{cnn}, Y^u_{tra}) \\
%  \mathcal{L}_{cps (tra)} &= \mathcal{L}_{dice}( P^u_{tra}, Y^u_{cnn}) 
% \end{align}
%
\begin{equation}
\small
\label{CPS}
 \mathcal{L}_{cps (cnn)} = \mathcal{L}_{dice}( P^u_{cnn}, Y^u_{tra}), \;\;\;
 \mathcal{L}_{cps (tra)} = \mathcal{L}_{dice}( P^u_{tra}, Y^u_{cnn}). 
\end{equation}
Here $\mathcal{L}_{dice}$ is the standard Dice loss function, but using pseudo-labels instead of ground-truth segmentation.
Note that during training there is no gradient back-propagation between $P^u_{cnn}$ and $Y^u_{cnn}$, and similar from $P^u_{tra}$ to $Y^u_{tra}$.

\subsection{Multi-Scale Cross Supervised Contrastive Learning (MCSC)}
\label{subsec:cscl}

Cross pseudo supervision 
%enables cross-teaching between the output layers of the two segmentation networks, on a single slice. However, it 
does not exploit feature regularities across the whole dataset, e.g. similarity between representations of the same organ in different slices.
We therefore add a contrastive loss, operating on multi-scale features extracted from the Transformer and the CNN. This has two advantages:
(i) It encourages consistency of the two models' internal features (not just outputs)
%from representation building level to better combine local spatial context modeling from convolutions in the CNN and global correlation modeling from long distance self-attention in the Transformer.
(ii) It captures high-level semantic relationships between distant regions, and between features on both labelled and unlabelled data.
%Cross pseudo supervision improves the model’s discriminative ability of a single image slice from the unlabeled data.Therefore, we introduce MCSC to model the high-level semantic relationships between distance regions between features in inter-slice and intra-slice way on both labeled and unlabeled data.

Our MCSC module (Figure~\ref{fig:contrastive}) is based on local supervised contrastive learning \cite{Wang2021a}, which learns a compact feature space by reducing the distance in the embedding space between positive pairs, and increasing the distance between negative pairs.
%) and the positive sample in the embedding space and pushing anchors away from negative samples. \pmh{do you actually have explicit anchors in your model? I thought just every pixel has a set of positives (and all others as negative)}
%\pmh{maybe add: However, in our case, instead of true supervision, we use the pseudo-labels from each network}
Firstly, it extracts features from the CNN and Transformer, then projects them into a common embedding space. This is followed by a novel approach of selecting positive and negative pairs using the pseudo labels, and a class-balanced contrastive loss calculated on these.

\paragraph{Feature Embedding.}
After $X= \{x_{i}\}_{i=1 \cdots N}$ is passed into $E_{cnn}(\cdot)$ and $E_{tra}(\cdot)$ respectively, the resulting features are projected by passing them through projectors $H_\mathrm{cnn}(\cdot)$ and $H_\mathrm{tra}(\cdot)$ into a unified feature space, where we will sample pairs to contrast.
Overall, we get a feature batch $F$ consisting of $2N$ feature maps $f_{i} = H(E(x_{i}))\in \mathbb{R}^{h\times  w\times c}$, where $f_{1\ldots N} $ come from the CNN and $f_{N+1\ldots 2N}$ from the Transformer (middle of Figure \ref{fig:contrastive}).
%\pmh{ok -- is it not simpler to name the features f\_cnn and f\_tra, each of size N, rather than a single thing with 2N elements?} 

%\vspace{-1em} \pmh{don't vspace it yet! better to do at the last moment when everything is written; also we can just redefine \paragraph}
\paragraph{Cross Supervised Sampling.}
For cross supervised sampling, we follow these strategies:
(i) We exchange class information from two models to guide the sampling, using the prediction of Transformer to be the supervisory information for CNN and vice-versa (Figure \ref{fig:contrastive}, right). This is consistent with the cross-prediction loss $\mathcal{L}_{cps}$, and implicitly it also makes the features predicted by the two models on the same slice consistent.
(ii) We contrast features on both unlabelled and labelled data. Since the pseudo labels are of varying quality, labelled data is included in the contrastive loss to reduce the noise.
(iii) We contrast pixels both within and between slices. Previous work focuses on inter-slice samples and ignores useful anatomical information within slices. For example, compared to different slices, the features of different class of organ boundaries in the image should be more similar. By focusing on them, we can refine the details of the hardest boundary segmentation.
Therefore, our strategy differs significantly from existing approaches to sampling pairs in supervised contrastive learning with semi-supervised segmentation, where positive or negative pairs are selected based on pseudo labels on unlabelled data \cite{zhao2023rcps, Wu2022Cross, Chaitanya2023}.

The computational complexity and memory for the supervised contrastive loss is very high; 
%, so existing works subsample a smaller set of pixel coordinates as positive pairs to fit in GPU memory \cite{Chaitanya2023}. However,
however, comparing many samples is crucial for improving the performance of contrastive learning \cite{He2020}. % Specifically, the overall computation complexity for the supervised local loss is $O(h^4)$, where $h$ is the size of an image, 256 in our case, which would necessitate $O(10^8)$ multiplications (\pmh{a gpu can do about $10^13$ ops/second... so this is not quite convincing}). 
To address this problem, inspired by \cite{Hu2021}, we compute the local contrastive loss over patches. We divide all the feature maps in $F$ into patches with size of $h' \times h'$. Let us assume there are $M$ patches of each $f$. We randomly select (without replacement) a patch from each feature map in $F$, and finally we get $M$ batches of $2N$ patches.
The loss is evaluated on $2N$ patches from each batch in turn, until the entire $f$ has been traversed. 
% \pmh{maybe add a sentence sayingt this reduces computational comleplexity from $O(h^4)$ to $O(M h'^2)$ or whatever it is}

\paragraph{Balanced Supervised Local Contrastive Loss.}
After sampling positive/negative pairs of pixels, { a contrastive loss is introduced to pull positive pairs closer and push negative pairs apart within the $2N$ patches. Given the extreme imbalance between background and foreground (different organs), a randomly sampled batch tends to consist of a significantly larger number of positive and negative pairs for the background, compared to the foreground organs. This inherent imbalance inevitably biases conventional supervised contrastive learning towards the background, consequently neglecting the differentiation of foreground categories. Simply eliminating the background during contrastive learning \cite{Hu2021} is not an optimal solution, as (i) the remaining number of foreground pixels is extremely small, and (ii) this fails to capture the relationship between the background and the foreground. 

Inspired by \cite{Zhu2022}, we average both the inter-class (positive) and intra-class (negative) feature contrast within the pixels of each class, and then forward it to calculate the supervised contrastive loss. In this way, each class makes an approximately balanced contribution. This balanced contrastive loss is implemented as follows:
\begin{equation}
\small
\begin{aligned}
\mathcal{L}_{bcl} = - \frac1{\lvert A\rvert} \sum_{a
_{i} \in A} \frac1{\lvert A_y\rvert -1 }\sum\limits_{p \in {A_y \backslash\{ i\}}} \mathrm{log} \frac {\mathrm{exp}(a_{i} \cdot a_p / \tau )}{\sum_{j \in {Y_A }}  \frac1{\lvert A_j\rvert}  \sum\limits_{a_k \in {A_j}} \mathrm{exp}(a_{i} \cdot a_k / \tau )},
\end{aligned}
\label{out}
\vspace{-5pt}
\end{equation}
where $A$ is the pixel-level feature sets of the $2N$ patches, $a_i$ represents the $i^{th}$ feature, $A_y$ is a subset that contains all samples of class $y$, $A_y\backslash\{ i\}$ represents all the pixels in $A_y$ excluding $a_i$, $Y_A$ represents the set of all the unique classes in current $A$, and $\tau$ is a temperature constant. By balancing the contribution of each class during contrastive learning, we avoid the learned representations being biased towards the dominant background.
Note that $\mathcal{L}_{bcl}$ is calculated over each $2N$ patches, and then averaged over $M$ batches of $2N$ patches for back-propagation. 
}

\paragraph{Multi-Scale Contrastive Loss.}
%
%Most \pmh{is there some that doesn't?}\textcolor{red}{[i think we are first one to do multi scale.]} 
Existing works on local contrastive learning pass the features of the last layer before the classifier into the projector.
However, the feature maps from earlier layers focus on coarser geometric information like the shape of organs, and later feature maps on details; both are important for segmentation, which depends both on relationships among multiple organs and gross anatomic structure (global) and textures of the specific tissues (local).
%However, this local contrast is easily prone to over-fitting on small medical datasets, and lacks generality \pmh{what kind of generality?}.
% Generally, small feature maps from early layers of decoder have large receptive fields  
%
We therefore pass features with $n$ different scales from $n$ layers of extractors and separate projectors, and then calculate each scale balanced contrastive loss $\mathcal{L}_{bcl}$ as $\mathcal{L}_{cl\_i}$. The overall loss $\mathcal{L}_{cl}$ is given by summing over each scale loss:
$\mathcal{L}_{cl} = (\mathcal{L}_{cl\_1} + \ldots+ \mathcal{L}_{cl\_n})$.

\subsection{Optimization}
The two networks are trained to minimize a weighted sum of the losses described in the previous sections:
$\scriptsize \mathcal{L}_{cnn} = \mathcal{L}_{sup(cnn)} + w_{cps}\mathcal{L}_{cps (cnn)} + w_{cl}\mathcal{L}_{cl}$
and $\scriptsize \mathcal{L}_{tra} = \mathcal{L}_{sup(tra)} + w_{cps}\mathcal{L}_{cps (tra)} + w_{cl}\mathcal{L}_{cl}$,
where $w_*$ are weighting factors used to balance the impact of individual loss terms. $w_{cps}$ is defined by a Gaussian warming up function \cite{Luo2022}: $w_{cps}(t_i) = 0.1\cdot e^{(-5(1-t_i/t_{total})^2)}$, where $t_i$ is $i$\textsuperscript{th} iteration of training and $t_{total}$ is the total number of iterations, while $w_{cl}$ is set to a constant value of $10^{-3}$ based on performance of the validation.
Note that the Transformer is used only during training, and does not contribute to the final inference -- the CNN is less computationally expensive, but has distilled the Transformer's knowledge.

\vspace{-5pt}
\section{Experiments}
\vspace{-5pt}

  \begin{figure}
  \vspace{-5pt}
    \centering
    \includegraphics[width=0.97\linewidth]{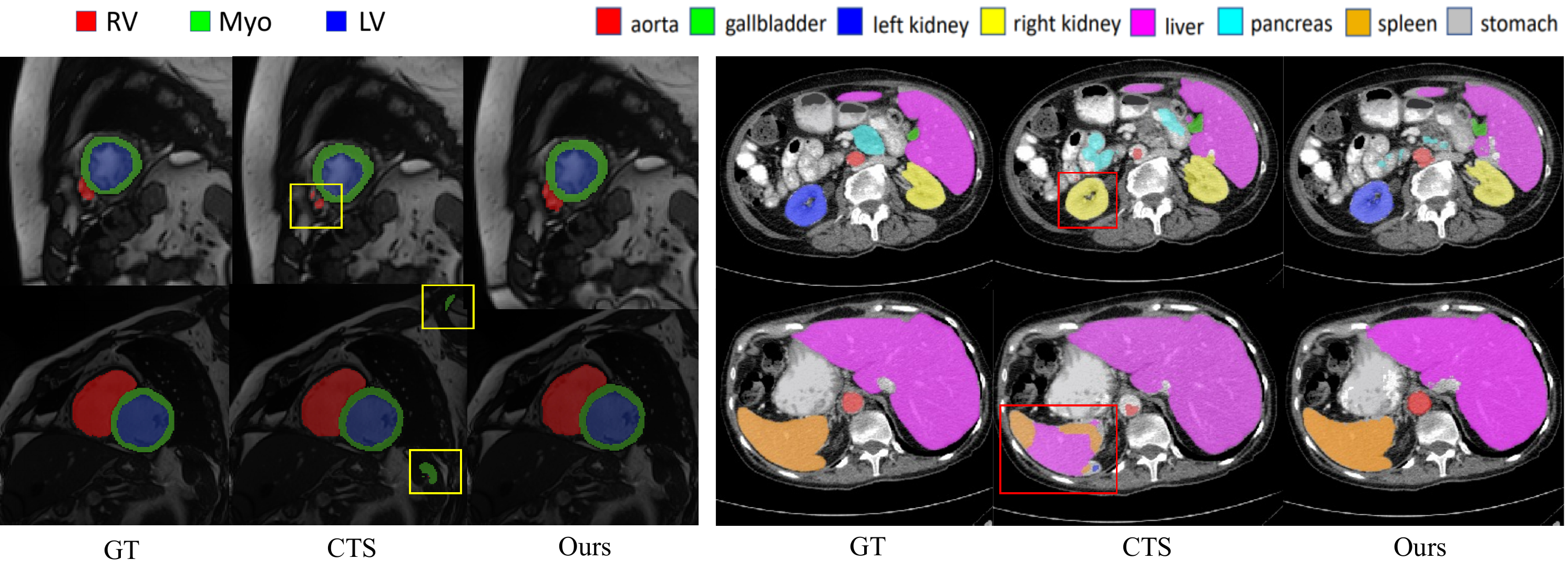} 
    \vspace{-8 pt}
    \caption{Qualitative results from our method and the best baseline CTS \cite{Luo2022} trained on 4 and 7 labelled cases on ACDC (\textbf{left}) and Synapse (\textbf{right}), respectively. }
    \label{fig:Visualacdc}
  \end{figure}

% \paragraph{Datasets and Evaluation Metrics.}
%
We evaluate our method on two benchmark datasets, ACDC \cite{bernard2018deep} and Synapse \cite{landman2015miccai}. 
ACDC contains 200 short-axis cardiac MR images from 100 cases (i.e. patients) with masks of the left ventricle (LV), myocardium (Myo), and right ventricle (RV) to be segmented; we follow the 
%Segmentation can further generate clinical measurements such as ejection fraction and end-diastolic volume. \pmh{no point saying this if you didn't actually calculate them!}
%Following the 
data split and the selection of labelled cases in \cite{Luo2022}.
%, the dataset is split into 70, 10 and 20 patients for training, validation and testing, respectively. 
Synapse contains abdominal CT scans from 30 cases with eight organs including aorta, gallbladder, spleen, left kidney, right kidney, liver, pancreas and stomach; the splits follow \cite{Chen2021Transunet}.
%, we used 18 cases (2212 slices) for training, and the remaining 12 cases for testing.
To quantitatively assess performance, we report two popular metrics: Dice coefficient (DSC) and $95\%$ Hausdorff Distance (HD).

We implemented our method in PyTorch. We used simple data augmentations to reduce overfitting: random cropping with a $224\times224$ patch, random flipping and rotations. 
All methods were trained till validation-set convergence (which was by 40,000 iterations). We selected the best checkpoint for evaluation based on validation set performance.
    Our method was trained using AdamW \cite{loshchilov2017decoupled} with a weight decay of $5\times10^{-4}$. We utilized the poly learning rate schedule, initialized at $5\times10^{-4}$ for CNN and $1\times10^{-4}$ for Transformer. The batch sizes were 4 and 10 respectively, with half labeled and half unlabeled images. 
For our MCSC module, each projector $H_*$ has two linear layers, where the first linear layer changes the dimension of feature map to 256 channels; the last layer has 128 channels and shares its parameters between the two models. In Eq.(2), temperature $\tau = 0.1$.
We use multi-scale feature maps from three layers of $E_*$, with sizes of $256\times256$, $56\times56$, and $28\times28$ respectively, and the size $h'$ of a patch was set to 19, 28 and 14 accordingly. All experiments were run on one (for ACDC) or two (for Synapse) RTX 3090 GPUs.

\vspace{-5pt}
\subsection{Comparison with Other Semi-Supervised Methods}

We compare our proposed method to several recent SSL methods that use U-Net as backbone, including Mean Teacher (MT) \cite{tarvainen2017mean}, Deep Co-Training (DCT) \cite{qiao2018deep}, Uncertainty Aware Mean Teacher (UAMT) \cite{yu2019uncertainty}, Interpolation Consistency Training (ICT) \cite{verma2022interpolation},  Cross Consistency Training (CCT) \cite{ouali2020semi}, Cross Pseudo Supervision (CPS) \cite{Chen2021a}, and the state-of-the-art (SOTA) method Cross Teaching Supervision (CTS) \cite{Luo2022}. Results for the weaker methods MT, DCT and ICT are given in the supplementary material.
% Our proposed MCSC is most similar to CTS, which is the first work to consider the difference in learning paradigms between CNN (U-Net \cite{ronneberger2015u}) and Transformer (Swin-Unet \cite{cao2023swin}).
We also compare against a U-Net trained with full supervision (UNet-FS), and one trained only on the labelled subset of data (UNet-LS).
Finally we compare with the SOTA fully-supervised Transformer based methods BATFormer \cite{Lin2023} on ACDC, and nnFormer \cite{zhou2021nnformer} on Synapse.
We retrained all the semi-supervised baselines using their original settings (optimizer and batch size), and report whichever is better of our retrained model or the result quoted in \cite{Luo2022}.

\paragraph{Results on ACDC.} Table \ref{tab:acdc} shows evaluation results of MCSC and the best-performing baseline under three different levels of supervision (7, 3 and 1 labelled cases). 
Our MCSC method trained on $10\%$ of cases improves both DSC and HD metrics compared to previous best SSL methods by a significant margin (more than $3\%$ on DSC and 5mm on HD). 
More importantly, it achieves 2.3mm HD, significantly better than even the fully supervised U-Net and BATFormer, which achieve 4.0 and 8.0 respectively. It also demonstrates competitive DSC of 89.4 $\%$, compared with 91.7 $\%$ and 92.8 $\%$ of U-Net and BATFormer. 
In addition, MCSC performance is highly resilient to the reduction of labelled data from $10\%$ to $5\%$, outperforming the previous SOTA SSL methods by around $10\%$ on DSC. The improvement is even more profound for the minority and hardest class, RV, with performance gains of 14 $\%$ on DSC and 13.6mm on HD.  
%from 5.4 and 9.4 to 1.1 and 3.5 on Myo and LV respectively. 
%When using labelled 3 cases, it dramatically improves the best mean Dice of CTS from 65.6 to 73.6 by a large margin, merely closing the UNet-LS trained on 7 cases. In particular, for the minority and hardest class - RV, MCSC achieved overwhelming gains, including a 14 $\%$ improvement on DSC and a 13.6 mm reduction on HD.
%The continuous victories also can be seen in the 1 case setting, and we even observe that the less labeled data, the greater the gap between our method and other opponents.
Figure \ref{fig:Visualacdc} shows qualitative results from UNet-LS, CPS, CTS and our method. MCSC produces a more accurate segmentation, with fewer under-segmented regions on minority class- RV (top) and fewer false-positive (bottom).
Overall, results prove that MCSC improving the semantic segmentation capability on unbalanced and limited-annotated medical image dataset by a large margin.

\begin{table}
  \centering
  \renewcommand*{\arraystretch}{0.9}
  %\begin{minipage}{1\linewidth}
    \centering

    \scalebox{0.7}{
    \begin{tabular}{c|c|c c|c c| c c| c c }
    \toprule
    \multirow{2}{*}{\textbf{Labelled}} & \multirow{2}{*}{\textbf{Methods}} & \multicolumn{2}{c|}{\textbf{Mean}}  & \multicolumn{2}{c|}{\textbf{Myo }}  & \multicolumn{2}{c|}{\textbf{LV }} & \multicolumn{2}{c}{\textbf{RV }} \\
    & & DSC$\uparrow$& HD$\downarrow$ & DSC$\uparrow$ & HD$\downarrow$ & DSC$\uparrow$ & HD$\downarrow$  & DSC$\uparrow$ & HD$\downarrow$ \\
    \midrule
   \multirow{2}{*}{ 70 cases (100$\%$)}
    & UNet-FS & 91.7  & 4.0  & 89.0  & 5.0 & 94.6  & 5.9  & 91.4 & 1.2  \\
    & BATFormer \cite{Lin2023} & 92.8  & 8.0    &90.26  & 6.8 & 96.30 & 5.9&  91.97  & 11.3  \\
     \hline
     \multirow{5}{*}{ 7 cases (10$\%$)}& UNet-LS  & 75.9   & 10.8 & 78.2  &8.6 & 85.5  &13.0& 63.9 &10.7  \\
    &CCT  \cite{ouali2020semi} & 84.0 & 6.6 & 82.3  & \underline{5.4} & 88.6& 
\underline{9.4} & 81.0 & 5.1 \\
    &CPS \cite{Chen2021a} & 85.0& \underline{6.6}& 82.9   &6.6 & 	88.0 &10.8 & 84.2 & \underline{2.3}\\
    &CTS  \cite{Luo2022} & \underline{86.4} & 8.6 & \underline{84.4} & 6.9  & \underline{90.1} & 11.2 & \underline{84.8} & 7.8\\
    & MCSC (Ours) & {\bf 89.4} &  {\bf2.3} &  {\bf87.6} & {\bf1.1}   & {\bf93.6} &  {\bf3.5} &  {\bf87.1} &   {\bf2.1}\\

     \midrule
     \multirow{5}{*}{ 3 cases (5$\%$)}& UNet-LS  & 51.2   & 31.2 & 54.8  & 24.4 & 61.8  &24.3& 37.0 & 44.4  \\
    &CCT  \cite{ouali2020semi}  & 58.6 & 27.9 & 64.7  & 22.4& 70.4& 27.1 & 40.8 &34.2 \\
    &CPS \cite{Chen2021a}   & 60.3& 25.5& 65.2   &18.3 & 	72.0 &22.2 & 43.8 & 35.8 \\
    &CTS \cite{Luo2022}   & \underline{65.6} & \underline{16.2} & \underline{62.8} & \underline{11.5}  & \underline{76.3} & \underline{15.7} & \underline{57.7} & \underline{21.4}\\
    & MCSC (Ours) & {\bf 73.6 } & {\bf 10.5 } & {\bf70.0} & {\bf8.8} & {\bf 79.2}&  {\bf14.9} & {\bf71.7} & {\bf7.8}\\
    \midrule
    \multirow{3}{*}{ 1 case }& UNet-LS  &  26.4   & 60.1 & 26.3  & 51.2 & 28.3  & 52.0& \underline{24.6} & \underline{77.0}  \\
    &CTS  \cite{Luo2022}  & \underline{46.8}  &  \underline{36.3}  & 55.1 &{\bf5.5}&	64.8 & {\bf4.1} &20.5 & 99.4\\
     &MCSC (Ours)  & {\bf58.6}  & {\bf31.2} &{\bf 64.2 }&\underline{13.3}&	{\bf78.1} &\underline{12.2} & {\bf33.5} &{\bf 68.1}\\
    \bottomrule 
    \multicolumn{10}{r}{\scriptsize \textbf{Best} is reported as bold, \underline{Second Best} is underlined. \hfill}
    \end{tabular}
    }
    \vspace{0 pt}
    \tabcaption{Segmentation results on DSC($\%$) and HD(mm) of our method and baselines on ACDC, across different numbers of labelled cases.}
    \vspace{-5 pt}
    \label{tab:acdc}

  \end{table}

\paragraph{Results on Synapse.}
Table \ref{tab:synapse} shows the segmentation results of the best-performing baselines on Synapse with 4 and 2 labelled cases. Compared to ACDC, Synapse is a more challenging segmentation benchmark as it includes a larger number of labelled regions with far more imbalanced volumes. Nevertheless, our method outperforms the baselines by a large margin. This demonstrates the robustness of our proposed framework, and the benefit of regularising multi-scale features from two models to be semantically consistent across the whole dataset. This is further highlighted in the qualitative results provided in Figure \ref{fig:Visualacdc}.

% \begin{figure}[h!]
% \begin{center}
% \includegraphics[width=\linewidth]{images/synapse+acdc.pdf}
%     \end{center}
% \vspace{-20pt}
% \caption{Visualizations from }
% \label{fig:synapse}
% \vspace{-10 pt}
% \end{figure}

\begin{table}[ht]

\centering
\scalebox{0.65}{
\begin{tabular}{c|c|c|c|c c c c c c c c}
\toprule
\textbf{Labelled}& \textbf{Methods} & DSC$\uparrow$ & HD$\downarrow$ & Aorta & Gallb & Kid$\_$L & Kid$\_$R& Liver& Pancr & Spleen& Stom\\
\midrule
\multirow{2}{*}{18 cases(100 $\%$)} &  UNet-FS  &75.6 & 42.3 & 88.8 & 56.1 & 78.9 & 72.6 & 91.9 & 55.8 & 85.8 & 74.7 \\
 &  nnFormer \cite{zhou2021nnformer} & 86.6  & 10.6 & 92.0 & 70.2 & 86.6 & 86.3 & 96.8 & 83.4 & 90.5 & 86.8 \\
\midrule
\multirow{5}{*}{ 4 cases(20 $\%$)} & UNet-LS  & 47.2 & 122.3 & 67.6 & 29.7 & 47.2 & 50.7 & 79.1 & 25.2 & 56.8 & 21.5 \\
 &CCT \cite{ouali2020semi} & 51.4 & 102.9 & 71.8 & 31.2 & 52.0 & 50.1 & 83.0 & 32.5 & 65.5 & 25.2 \\
 &CPS \cite{Chen2021a}   & 57.9 & 62.6 & \underline{75.6} & \bf{41.4} & 60.1 & 53.0 & \underline{88.2} & 26.2 & 69.6 & 48.9 \\
&CTS \cite{Luo2022} & \underline{64.0} & \underline{56.4} & \bf{79.9} & \underline{38.9} & \underline{66.3} & \underline{63.5} & 86.1 & \bf{41.9} & \underline{75.3} & \bf{60.4} \\

& MCSC (Ours) &  \bf{66.8} &  \bf{28.2} & 75.2 & 37.2 & \bf{73.1} & \bf{71.5} & \bf{92.4} & \underline{41.0} & \bf{86.1} & \underline{57.8} \\
\midrule
\multirow{5}{*}{ 2 cases(10 $\%$)} & UNet-LS  &45.2 & \underline{55.6} & 66.4 & \bf{27.2} & 46.0 & 48.0 & 82.6 & 18.2 & 39.9 & 33.4 \\
 &CCT \cite{ouali2020semi} & 46.9 & 58.2 & 66.0 & \underline{26.6} & 53.4 & 41.0 & 82.9 & 21.2 & 48.7 & 35.6 \\
 &CPS \cite{Chen2021a} & 48.8 & 65.6 & 70.9 & 21.3 & 58.0 & 45.1 & 80.7 &  \underline{23.5} &  \underline{58.0} & 32.7 \\
&CTS \cite{Luo2022} & \underline{52.0} & 63.7 & \bf{73.2} & 12.7 & \underline{67.2} & \underline{64.7} &  \underline{82.9} & \bf{31.7} & 40.9 & \bf{42.4} \\

& MCSC (Ours) &  \bf{55.8} & \bf{25.9} & \underline{72.6} & 20.4 & \bf{73.8} & \bf{71.3} & \bf{91.0} & 3.5 & \bf{75.4} & \underline{38.6} \\
\bottomrule
    \multicolumn{12}{r}{\scriptsize \textbf{Best} is reported as bold, \underline{Second Best} is underlined. \hfill}
\end{tabular}}

\caption{Comparison with different models on Synapse. The performance is reported by class-mean DSC (\%) and HD (mm), as well as the DSC value for each organ. }
\vspace{-5pt}
% Gallbladder, left Kidney, right Kidney, Pancreas and Stomach are abbreviated as Gallb, Kid$\_$L, Kid$\_$R, Pancr and Stom.}
\label{tab:synapse}
\end{table}

\subsection{Ablation study}

In Table \ref{tab:ablation} we explore the influence of proposed modules on the performance on ACDC with 7 labelled cases. 
Starting from CTS \cite{Luo2022} (top row), and adding supervised local contrastive learning (SCL) with a prior approach for balancing the loss (DB \cite{Hu2021}), we observe a significant improvement of $1.1\%$ on DSC; this emphasizes the importance of enforcing consistency between features of the two models.  % from representation building level and building high-level semantic relationships between distance regions between features for SSL. 
By exchanging class information from CNN and Transformer to select contrasted samples (instead of using each model's own predictions as pseudo-labels), 
% the consistency is implicitly introduced from feature building level and the whole framework (supervision information is exchanged in the same way for external $\mathcal{L}_{cps}$). It leads to
we see an improvement in DSC and HD from 87.50 to 88.23 and 7.4 to 3.4 respectively. 
% In order to better deal with class imbalance problem of medical image segmentation, instead of only discard dominant pixels of background in anchors, 
Our approach to balancing different classes (Balanced), instead of just discarding background pixels (DB), improves DSC by 0.7\%, since minority classes are better separated. 
Finally, utilizing multi-scale instead of just final-layer features further improves performance by 0.58\% and 2.3\% DSC and HD respectively. 

In Table \ref{tab:branches of muti-scale contrastive}, we compare results using different feature maps as input to the contrastive loss; we see best performance is achieved by using both $256\times256$ and $28\times28$ feature maps. 
Thus, combining coarser geometric information in global features and detailed local features does indeed benefit medical image segmentation. 

\begin{table}[ht]
  \renewcommand*{\arraystretch}{0.9}
\begin{minipage}[c]{0.62\textwidth}
\centering
\begin{center}
\scalebox{0.78}
{
\begin{tabular}{c c c c c |c|c}
\toprule
   SCL & DB & CroLab & Balanced & MulSca & DSC$\uparrow$ & HD$\downarrow$  \\

\midrule
 &  &  &  & &86.40 & 8.6 \\
\cmark & \cmark &  &  & & 87.50 & 7.4 \\
 \cmark & \cmark & \cmark &  & & 88.23 &  3.4 \\
\cmark&    & \cmark & \cmark&  &  88.80 &  4.6   \\
    \midrule
\cmark &   & \cmark & \cmark& \cmark &  {\bf 89.38} &  {\bf2.3}  \\
\bottomrule
\end{tabular}
}
\end{center}
\vspace{-6pt}
\caption{Ablation study for the primary components of our model. SCL denotes supervised local contrastive loss. DB denotes discarding background pixels as anchor. CroLab stands for cross label information of two models to select contrastive sample. Balanced means averaging the instances of each class in denominator of SCL. MulSca means contrasting multi-scale feature maps. }
\label{tab:ablation}
\end{minipage}
\hfill
\begin{minipage}[c]{0.32\textwidth}
\centering
\scalebox{0.7}{
    \begin{tabular}{ccc|c c}
    \toprule
    \multicolumn{3}{c|} {\textbf{Branches}} & \multicolumn{2}{c}{\textbf{Mean}}  \\  256 & 56 & 28 & DSC$\uparrow$ & HD$\downarrow$  \\
    \midrule
     \cmark & & & 88.80   &   4.6  \\
     & \cmark & & 88.88 & 4.2 \\
     & & \cmark & 88.39 & 4.5 \\
     \cmark & & \cmark & {\bf 89.38} &  {\bf2.3} \\
     \cmark  & \cmark & & 88.92   & 2.9 \\
     \cmark & \cmark & \cmark & 88.35 & 4.3 \\
 \bottomrule
    \end{tabular}
}

\vspace{3pt}
\caption{Ablation analysis on the choice of feature maps for the multi-scale contrastive loss (ACDC, 7 labelled cases). Full table is in the supplementary material. ~\\}
\label{tab:branches of muti-scale contrastive}
\end{minipage}

\end{table}

\subsection{Visualization of features with contrastive learning}

\begin{figure}[ht!]
\begin{center}
\includegraphics[width=\linewidth]{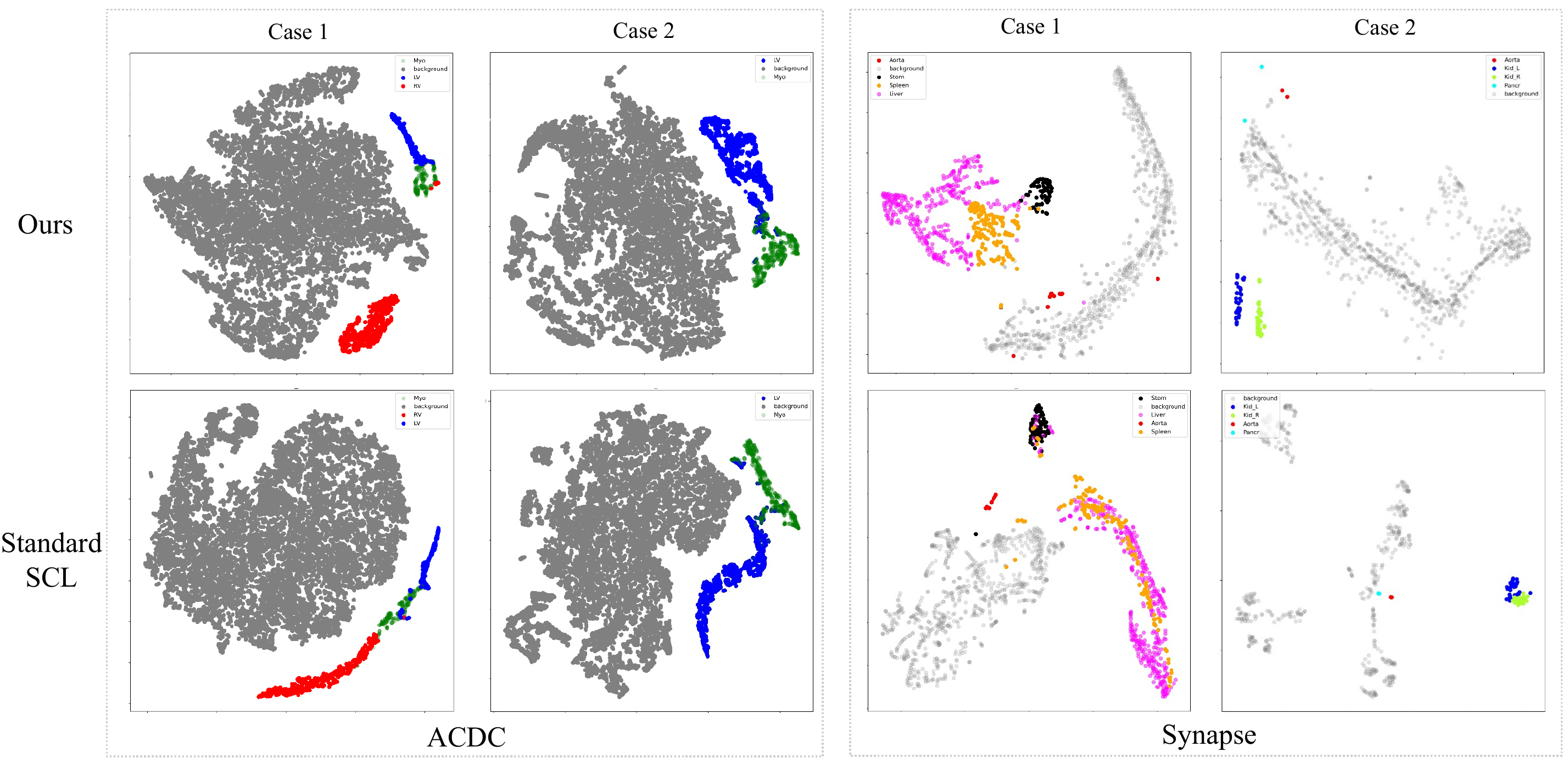}
    \end{center}
\vspace{-10pt}
\caption{Visualization of embedding features from our method and standard SCL after applying t-SNE on test data of ACDC and Synapse respectively. Each case represents one slice from a different patient.}
\label{fig:contrastive2}
\vspace{0 pt}
\end{figure}

Figure \ref{fig:contrastive2} shows visualizations of embedding features applying t-SNE on a single slice of two cases from the test data of ACDC and Synapse respectively. The models are trained with 7 cases (ACDC) and 4 cases (Synapse).
Different colors represent different classes. Features are taken from the feature map after the projector with scale of $256\times256$, and each point in the figure is the embedding of one pixel.
The standard SCL is the second row of Table 3 in the main text (SCL+DB). 
For ACDC, the left case 1 shows our method better separates RV from the other two foreground classes, and reduces the overlap between LV and Myo. For the case 2, the foreground clusters of ours are tighter.
A more clear and consistent effect can be seen on Synapse. For the case on the left, our method makes the liver, spleen and stomach much better separated than standard SCL. A similar situation also occurs with the left and right kidneys for the case 2.
Overall, through cross labelling, averaging the contribution of each class in SCL, and contrasting multi-scale feature maps, our method obtains a better embedding representations for segmentation, where features within the same class are pulled closer and features for different class are spread farther apart. 

\vspace{-18pt}
\section{Conclusion}
\vspace{-5pt}
% We have proposed cross pseudo supervision based on a unified, multi-scale, local contrastive learning approach. Our method accounts for the prevalence of different classes to alleviate the problem of imbalanced data. 
% We demonstrate that our proposed method significantly enhances semi-supervised learning for medical image segmentation. 
% Our results on two challenging benchmark segmentation tasks images show that our method closes the gap with state-of-the art fully supervised learning, and demonstrates remarkably resiliant performance even when the labelled data are significantly reduced. 
% Future work should aim to test our approach on clinical populations with significant anatomical variabilities. 
We have presented a novel SSL framework for medical image segmentation based on cross-teaching between a Transformer and a CNN.  
This incorporates a supervised local contrastive loss, named MCSC, that encourages intra-class feature similarity and inter-class discriminativity across the whole dataset.
%It uses a  to exploit consistency regularisation from feature building-level and prediction output-level. % this did not make sense, and is too long anyway
Furthermore, it addresses class imbalance with a loss that eliminates the negative effects of excessive background pixels.
Finally, it contrasts multi-scale feature maps, to combine global and local feature understanding.
Our experiments on two commonly used medical datasets demonstrate that the proposed framework can fully take advantage of labelled and unlabelled data, and demonstrates remarkably resiliant performance even when the labelled data are significantly reduced.
% \input{conclusions}
% \section{Acknowledgement }
% \pmh{not in review version!}

\section{Acknowledgements}
The authors acknowledge funding by China Scholarship Council, EPSRC (EP/W01212X/1) and Royal Society (RGS/R2/212199).

\printbibliography

\end{document}